\PassOptionsToPackage{unicode}{hyperref}
\PassOptionsToPackage{hyphens}{url}
\PassOptionsToPackage{dvipsnames,svgnames,x11names}{xcolor}
\documentclass[
  11pt,
]{article}
\usepackage{xcolor}
\usepackage[margin=1in]{geometry}
\usepackage{amsmath,amssymb}
\usepackage{iftex}
\ifPDFTeX
  \usepackage[T1]{fontenc}
  \usepackage[utf8]{inputenc}
  \usepackage{textcomp} 
\else 
  \usepackage{unicode-math} 
  \defaultfontfeatures{Scale=MatchLowercase}
  \defaultfontfeatures[\rmfamily]{Ligatures=TeX,Scale=1}
\fi
\usepackage{lmodern}
\ifPDFTeX\else
\fi
\IfFileExists{upquote.sty}{\usepackage{upquote}}{}
\IfFileExists{microtype.sty}{
  \usepackage[]{microtype}
  \UseMicrotypeSet[protrusion]{basicmath} 
}{}
\makeatletter
\@ifundefined{KOMAClassName}{
  \IfFileExists{parskip.sty}{%
    \usepackage{parskip}
  }{
    \setlength{\parindent}{0pt}
    \setlength{\parskip}{6pt plus 2pt minus 1pt}}
}{
  \KOMAoptions{parskip=half}}
\makeatother
\usepackage{longtable,booktabs,array}
\usepackage{calc} 
\usepackage{etoolbox}
\makeatletter
\patchcmd\longtable{\par}{\if@noskipsec\mbox{}\fi\par}{}{}
\makeatother
\IfFileExists{footnotehyper.sty}{\usepackage{footnotehyper}}{\usepackage{footnote}}
\makesavenoteenv{longtable}
\usepackage[round]{natbib}
\usepackage{bookmark}
\IfFileExists{xurl.sty}{\usepackage{xurl}}{} 
\hypersetup{
  pdftitle={Can Open LLMs Clear Sanctions-Screening Alerts? A Pre-registered Test on Public Sanctions Lists with Turkish Name Variants},
  colorlinks=true,
  linkcolor={Maroon},
  filecolor={Maroon},
  citecolor={Blue},
  urlcolor={Blue},
  pdfcreator={LaTeX via pandoc}}

\title{Can Open LLMs Clear Sanctions-Screening Alerts? A Pre-registered
Test on Public Sanctions Lists with Turkish Name Variants}
\author{Mehmet Murat Huyut\\
Kuveyt Türk Participation Bank and İstanbul Bilgi University, İstanbul,
Türkiye\\
mhuyut@gmail.com; ORCID 0000-0003-1812-6755}
\date{November 2026}

\begin{document}
\maketitle
\begin{abstract}
Sanctions screening raises name-based alerts that are mostly false
positives, and level-1 analysts clear them by checking the customer's
date of birth and name against a written policy. We test whether
open-weight language models that a bank can run on its own premises can
perform this level-1 disposition. From the UK and EU sanctions lists
(7,103 natural persons) we built 12,200 synthetic customers: listed
persons under six name conditions, including Turkish spellings of
Arabic-origin names and Turkish romanisations of Cyrillic names, and two
date-of-birth conditions, plus namesakes, near-name customers and a
general population. A deterministic fuzzy and embedding screen raised
7,383 alerts, which three open models served on domestic infrastructure
and three rule baselines dispositioned under the same policy. The models
sent far fewer innocent customers to level 2 than a rule that applies
the policy's date clause: 0.6 to 16.9 per 1,000 screened customers
against 24.4, and they stated the date-of-birth status correctly in
98.3\% to 98.5\% of alerts. They also failed to escalate 1.5\% to 2.2\%
of the listed persons. The pre-registered non-inferiority test against
the rule (margin 2 percentage points) was met, but only because of a
construction error we found after the analysis: 21 customers in the
``minor discrepancy'' condition had dates that conflict under the
policy. Without them the rule misses no listed person and no model meets
the margin (upper confidence bounds +2.5 to +3.2 points). The misses had
model-specific causes: one model read minor date discrepancies as
conflicts, one declared word-order and typing variants to be different
names, and one discounted alerts that its own rationale said to
escalate. Fuzzy matching found 69\% of the Turkish spellings of
Arabic-origin names; adding embeddings raised this to 74\%. Open LLMs
can cut level-1 workload, but on this evidence only behind a
deterministic date check and with every name doubt escalated.
\end{abstract}

\section{Introduction}\label{introduction}

Every bank screens its customers and their payments against sanctions
lists. Screening is a name search, and a name search over hundreds of
thousands of customers against thousands of listed persons, many with
common given names, transliterated surnames and several aliases,
produces alerts that are almost all false positives. A level-1 analyst
clears each alert by comparing the customer with the list entry: does
the date of birth conflict with every date the list records, and is the
name really a different name or only a different spelling of the same
one? Alerts that cannot be cleared go to a level-2 investigator. The
level-1 step is repetitive, rule-bound and expensive, which makes it an
obvious candidate for a language model.

It is also a step where a wrong ``discount'' has a direct legal cost: a
listed person who is waved through is a sanctions breach. FATF
Recommendations 6 and 7 require targeted financial sanctions to be
applied without delay \citep{fatf2012rec}, and Türkiye implements the
corresponding UN Security Council sanctions, with freezing ``without
delay'' (\emph{gecikmeksizin}), through Laws 6415 and 7262
\citep{tr6415, tr7262}. The EU's new anti-money-laundering regulation,
which applies from July 2027, permits due diligence decisions resulting
from automated processes or AI systems only if decisions to enter into,
refuse or maintain a business relationship, or to increase or reduce due
diligence, are ``subject to meaningful human intervention'', and the
customer can obtain an explanation and challenge the decision
\citep[Art. 76(5)]{eu2024amlr}. In Türkiye, banks must keep their
primary and secondary information systems in the country, including
outsourced and cloud services \citep{bddk2020bs}, and personal data may
be transferred abroad only under restrictive conditions \citep{tr6698}.
A bank that wants to use a language model on customer data therefore
needs a model it can run itself, which in practice means an open-weight
model on its own or domestic infrastructure.

Whether such models can clear sanctions alerts safely has not been
tested in public. Work on LLMs for sanctions and know-your-customer
tasks either uses proprietary or in-house data
\citep{he2025aria, chourasia2026sger, wahib2026spectra} or evaluates
pairwise entity matching, where the task is to decide whether two
records describe the same entity \citep{smith2026opensanctions}. Level-1
disposition is a different task: the model is given a bank policy and
must decide whether there is \emph{clear evidence} that the customer is
not the listed person, and the costs of its two errors are very unequal.

We ran a pre-registered experiment on public data. The watch list is the
natural persons on the UK and EU sanctions lists. The customers are
synthetic: listed persons whose names are rendered the way they would
appear in a Turkish bank (Turkish spellings of Arabic-origin names,
Turkish romanisations of Cyrillic names, word-order changes, typing
errors with Turkish upper case), with dates of birth that are either
compatible with the list or differ by a minor discrepancy the policy
tolerates; namesakes whose dates conflict; near-name customers with a
different family name; and a general population. A deterministic screen
raises the alerts, and three open models and three rule baselines
disposition them under one written policy. We measure the two quantities
a compliance officer would ask about first, the share of listed persons
not escalated and the number of innocent customers sent to level 2, and
we check the stated reasons against the facts.

Our contributions are: (i) a reproducible benchmark for level-1
sanctions disposition built only from public lists and synthetic
customers, with a Turkish name stratum; (ii) a pre-registered comparison
of open models served on domestic infrastructure with rule baselines
under one policy, reporting misses and level-2 workload together; and
(iii) an account of a construction error that reversed our safety
conclusion, which we report next to the registered result.

\section{Related work}\label{related-work}

\textbf{Name matching and sanctions screening.} Record linkage has a
long statistical tradition
\citep{fellegi1969theory, jaro1989advances, christen2012datamatching},
and name matching in particular depends on the choice of string
comparator \citep{bilenko2003adaptive} and on transliteration across
scripts \citep{pouliquen2005multilingual}. LLMs have been applied to
entity resolution to improve or cheapen matching
\citep{li2024booster, zeakis2026distiller}. The closest public benchmark
is OpenSanctions Pairs \citep{smith2026opensanctions}, which releases
expert-labelled pairs over sanctions and watchlist entities and reports
that LLMs, including a locally deployable distilled model, outperform a
rule matcher, with LLM errors concentrated in cross-script
transliteration and in treating minor discrepancies, such as off-by-one
dates, as evidence of a non-match. Cross-script failures have also been
diagnosed for Indic scripts, where same-person accuracy collapses while
the model's scores still rank matching pairs well above non-matching
ones; much, though not all, of the gap is a misplaced decision threshold
\citep{gopal2026crossscript}. For corporate-family resolution, where
linked records are deliberately different entities, blocking rather than
pairwise matching is the binding constraint \citep{gupta2026corpfam}. We
know of no study of Turkish name forms in screening.

\textbf{LLMs in KYC and AML.} Production studies report LLM pipelines
for KYC document processing \citep{wahib2026spectra}, customer due
diligence name screening with test-time rule updates \citep{he2025aria}
and fine-tuned name matching on identity data \citep{chourasia2026sger},
all on proprietary data. Other work uses LLM agents for adverse-media
screening \citep{chernakov2025adverse}, drafting suspicious activity
narratives \citep{naik2025coinvestigator} and triaging AML alerts under
an evidence contract on public synthetic benchmarks
\citep{torres2026triage}; a serving study addresses self-hosting open
models for AML workloads \citep{naik2026llmops}. Before LLMs,
machine-learning triage on top of rule alerts cut false positives
substantially while keeping most true positives
\citep{nasereddin2021alert}. The closest task is ARIA's adjudication of
watchlist hits, which reports both sensitivity and specificity, but on
proprietary data and without a fixed written policy \citep{he2025aria}.
We know of no study that tests level-1 disposition of sanctions alerts
against a written policy on public data with both error types measured.

\textbf{Pre-registration.} We follow proposals to pre-register NLP
experiments \citep{vanmiltenburg2021prereg} and are aware of the
arguments against it \citep{sogaard2023prereg}; our deviation section
shows one reason it is useful.

\section{Method}\label{method}

\subsection{Watch list}\label{watch-list}

We used the UK Sanctions List \citep{uk_sanctions_list} and the EU
consolidated financial sanctions list \citep{eu_fsf_list}, both
downloaded on 4 November 2026. We kept natural persons with at least one
recorded date of birth and at least one Latin-script name: 7,103 entries
(3,330 UK, 3,773 EU), of which 5,090 also carry names in a non-Latin
script. Entries of the same person on both lists were linked when they
shared a normalised name and a date of birth, giving 5,296 persons. We
had planned to include the US SDN list \citep{ofac_sdn}, but its
download server could not be resolved from our machine on the
registration date, so it is not used.

\subsection{Synthetic customers}\label{synthetic-customers}

All customers are synthetic; no bank or customer data were used. Pilot
and full sets are drawn from disjoint persons. Positive customers are
listed persons under six name conditions crossed with two date-of-birth
conditions. The name conditions are the exact primary name; a
good-quality alias; a Turkish spelling of an Arabic-origin name (for
example Abd al-Rahman as Abdurrahman); a Turkish romanisation of a
Cyrillic name; a change of word order or a dropped middle name or
patronymic; and one or two typing errors with Turkish upper case. The
date of birth either agrees with the list on every component the list
records, or differs by a minor discrepancy that the policy says is not a
conflict (day and month transposed, year off by one with the same day
and month, or one digit of the day changed). Negative customers are
namesakes (the same six name conditions with a date of birth at least
five years from every listed date), near-name customers (the family name
replaced by a different, similar-looking family name, same date of
birth) and a general population of common Turkish names and recombined
list names with random dates. The full set has 1,200 positive and 11,000
negative customers (12,200 in total; the pilot used 2,220).

\subsection{Stage 1: screening}\label{stage-1-screening}

Screening is deterministic. Channel F computes RapidFuzz token-set
similarity \citep{bachmann2026rapidfuzz} between the normalised customer
name and every Latin name of an entry (Turkish-aware lower case, ASCII
folding, punctuation removed) and raises an alert for each entry scoring
at least 85, at most 5 per customer. Channel E embeds all names,
including non-Latin ones, with BGE-M3 \citep{chen2024bgem3} run locally,
and adds the entries in a customer's top 3 by cosine similarity that
reach a threshold \(\tau\) and are not already F alerts. \(\tau=0.79\)
was fixed on the pilot by a registered rule (the smallest value at which
general-population pilot customers received on average at most 0.10
E-only alerts). Pipeline F uses channel F alone; pipeline F+E uses both.
In the full set pipeline F raised 5,823 alerts and F+E 7,383.

\subsection{Stage 2: level-1
disposition}\label{stage-2-level-1-disposition}

The bank policy (291 words) allows a discount only on clear evidence
that the customer is not the listed person: a date of birth incompatible
with every listed date, with minor discrepancies explicitly not counted
as conflicts, or a name that is a different name from every name and
alias on the entry, with spelling, transliteration, word-order and
dropped-part variants explicitly counted as the same name. A nationality
difference alone is not grounds to discount. Otherwise the alert is
escalated.

Three rules dispositioned every alert. R0 escalates everything (no
level-1 step). R1 escalates only an exact date-of-birth match. R2
applies the policy's date clause exactly and represents the name clause
by the fuzzy score: it discounts a date conflict, and otherwise
escalates if the F score reaches the alert threshold. In pipeline F
every alert reaches that threshold, so R2 there discounts only date
conflicts.

The models were deepseek-v4.1-flash, qwen3.8-flash-next and gemma-4-31b,
open-weight models served by EVREN, a domestic (Türkiye-hosted) research
inference API, at temperature 0 with a fixed seed. Each alert was sent
separately with the policy, the customer record (name, date of birth,
nationality) and the full list entry (all names with their type and
script, dates of birth, nationalities, country of birth). The model
returned JSON with a name assessment (same, variant, different), a
date-of-birth assessment (match, minor discrepancy, conflict), a
decision and a reason. A customer is escalated if any of its alerts is
escalated. Unparseable replies count as escalate, the safe default of a
level-1 queue.

\subsection{Ground truth}\label{ground-truth}

A positive customer is correctly dispositioned if escalated, provided
stage 1 raised an alert on the person's own entry; otherwise the miss is
attributed to stage 1. A negative customer is correctly dispositioned if
every alert is discounted, provided each alert meets a policy discount
condition by construction (a date conflict, or for near-name customers a
family name that resembles no name part of the entry). Negative
customers with an alert that meets neither condition are excluded as
indeterminate (72 in pipeline F).

\subsection{Pre-registration, pilot, amendments and a
deviation}\label{pre-registration-pilot-amendments-and-a-deviation}

Hypotheses, outcomes, tests and gates were written and sealed with
SHA-256 hashes before any model call on study data. Two amendments were
made before the full run. The first withdrew a planned comparison of
local and domestic serving of the same model: the deployability claim
rests on the models being open weights, not on where we ran them. The
second applied the pilot gate (at least 90\% parseable decisions per
model, and fuzzy recall on exact names of at least 95\%): fuzzy recall
on exact names was 100\%, and the three models above returned 96.2\%,
100.0\% and 100.0\% parseable decisions, while glm-5.3, piloted on a
subsample of 150 alerts, returned 59.3\% (its hidden reasoning used the
whole output budget) and was dropped. One local model, qwen3.8:27b,
completed the pilot before the first amendment; we report it only as
exploratory.

After the confirmatory analysis we traced every miss of R2, which should
miss no positive in pipeline F because no positive was built with a
conflicting date. All of them came from one generator branch: when
changing one digit of the day produced day 0 or a day above 28, the
generator set the day to 18, which for list days 28 to 31 is not a minor
discrepancy but a conflict. 21 of the 600 positives in the
minor-discrepancy condition of the full set are affected (2 in the
pilot, which we did not notice at the gate). We did not regenerate data
or rerun any model. We report the registered analysis as registered and,
next to it, a post hoc sensitivity analysis that excludes the affected
customers. The deviation is logged and hashed with the pre-registration.

\subsection{Analysis}\label{analysis}

Rates have Wilson 95\% intervals \citep{wilson1927probable}. H1 (safety)
is a non-inferiority test of each model's miss rate in pipeline F
against R2 with a margin of 2 percentage points, using the paired
Newcombe interval for the difference \citep{newcombe1998paired}. H2
(workload) compares escalated negatives with R2 in pipeline F; H4
compares escalated positives with R2 in pipeline F+E; both use one-sided
exact McNemar tests. H3 compares fuzzy recall for the two
Turkish-transliteration conditions with exact names (Fisher) and tests
whether channel E raises it (McNemar). H5 compares misses under minor
discrepancies with compatible dates (Fisher). H6 tests whether the
stated date-of-birth assessment agrees with the deterministic status in
more than 95\% of alerts (exact binomial). Within each hypothesis we
apply Holm's correction across models \citep{holm1979simple}.
Confirmatory tests use the full set only.

\section{Results}\label{results}

\subsection{Stage 1}\label{stage-1}

\begin{longtable}[]{@{}lrrr@{}}
\caption{Stage-1 recall of positive customers (alert on the person's own
entry), full set.}\tabularnewline
\toprule\noalign{}
Name condition & \(n\) & Pipeline F & Pipeline F+E \\
\midrule\noalign{}
\endfirsthead
\toprule\noalign{}
Name condition & \(n\) & Pipeline F & Pipeline F+E \\
\midrule\noalign{}
\endhead
\bottomrule\noalign{}
\endlastfoot
Exact primary name & 200 & 100.0\% & 100.0\% \\
Alias & 200 & 100.0\% & 100.0\% \\
Turkish spelling of Arabic-origin name & 200 & 69.0\% & 74.0\% \\
Turkish romanisation of Cyrillic name & 200 & 90.5\% & 96.0\% \\
Word order / dropped part & 200 & 100.0\% & 100.0\% \\
Typing errors & 200 & 99.0\% & 99.5\% \\
All & 1200 & 93.1\% & 94.9\% \\
\end{longtable}

Fuzzy matching found every exact, alias and word-order positive but only
69.0\% of the Turkish spellings of Arabic-origin names and 90.5\% of the
Turkish romanisations of Cyrillic names (Table 1). Embeddings closed a
larger share of the gap for Cyrillic names, where the entry often
carries the original script, than for Arabic-origin names. A listed
person missed at stage 1 never reaches the level-1 analyst, human or
model, so these are the largest misses in the system.

\subsection{Misses and level-2
workload}\label{misses-and-level-2-workload}

\begin{longtable}[]{@{}llrlrr@{}}
\caption{Level-1 disposition, full set. Miss: share of positive
customers with an alert on their own entry who were not escalated. FP:
share of alerted negative customers escalated. Workload: escalated
negative customers per 1,000 screened customers. Registered analysis
(all customers).}\tabularnewline
\toprule\noalign{}
Pipeline & Method & Miss & 95\% CI & FP & Workload \\
\midrule\noalign{}
\endfirsthead
\toprule\noalign{}
Pipeline & Method & Miss & 95\% CI & FP & Workload \\
\midrule\noalign{}
\endhead
\bottomrule\noalign{}
\endlastfoot
F & R0 escalate all & 0.0\% & 0.0\%--0.3\% & 100.0\% & 135.5 \\
F & R1 strict date rule & 49.1\% & 46.1\%--52.0\% & 18.0\% & 24.4 \\
F & R2 policy rule & 1.8\% & 1.2\%--2.8\% & 18.0\% & 24.4 \\
F & deepseek-v4.1-flash & 1.9\% & 1.2\%--2.9\% & 12.5\% & 16.9 \\
F & qwen3.8-flash-next & 2.2\% & 1.5\%--3.3\% & 5.2\% & 7.0 \\
F & gemma-4-31b & 1.5\% & 0.9\%--2.4\% & 0.4\% & 0.6 \\
F+E & R0 escalate all & 0.0\% & 0.0\%--0.3\% & 100.0\% & 176.6 \\
F+E & R1 strict date rule & 49.1\% & 46.2\%--52.0\% & 14.3\% & 25.2 \\
F+E & R2 policy rule & 3.5\% & 2.6\%--4.8\% & 13.8\% & 24.3 \\
F+E & deepseek-v4.1-flash & 1.8\% & 1.2\%--2.8\% & 11.9\% & 21.0 \\
F+E & qwen3.8-flash-next & 2.3\% & 1.6\%--3.3\% & 4.6\% & 8.1 \\
F+E & gemma-4-31b & 1.7\% & 1.1\%--2.6\% & 0.3\% & 0.6 \\
\end{longtable}

All three models sent fewer innocent customers to level 2 than R2 (Table
2). In pipeline F, gemma-4-31b escalated 7 of the 1,653 alerted negative
customers against 298 for R2, a reduction of 98\%; qwen3.8-flash-next
cut the workload by 71\% and deepseek-v4.1-flash by 31\%. The models
used the name clause, which R2 cannot: near-name customers share the
listed person's date of birth and differ in the family name, so only a
judgment about the name can clear them. R1, which demands an exact date,
missed about half of the listed persons, because half of them were built
with a minor discrepancy; it shows the cost of a rule that ignores the
policy's tolerance.

In pipeline F+E, R2 missed 3.5\% of positives: it discounts
embedding-only alerts below the fuzzy threshold, which include Turkish
spellings that only the embedding channel found. The models judged those
alerts on their merits.

\subsection{Hypothesis tests}\label{hypothesis-tests}

\begin{longtable}[]{@{}
  >{\raggedright\arraybackslash}p{(\linewidth - 10\tabcolsep) * \real{0.1667}}
  >{\raggedright\arraybackslash}p{(\linewidth - 10\tabcolsep) * \real{0.1667}}
  >{\raggedright\arraybackslash}p{(\linewidth - 10\tabcolsep) * \real{0.1667}}
  >{\raggedleft\arraybackslash}p{(\linewidth - 10\tabcolsep) * \real{0.1667}}
  >{\centering\arraybackslash}p{(\linewidth - 10\tabcolsep) * \real{0.1667}}
  >{\centering\arraybackslash}p{(\linewidth - 10\tabcolsep) * \real{0.1667}}@{}}
\caption{Pre-registered tests (Holm-adjusted, \(\alpha=0.05\); H1
one-sided at 0.025). Registered: all customers. Sensitivity: excluding
the 21 positives affected by the construction error (post
hoc).}\tabularnewline
\toprule\noalign{}
\begin{minipage}[b]{\linewidth}\raggedright
Hypothesis
\end{minipage} & \begin{minipage}[b]{\linewidth}\raggedright
Model
\end{minipage} & \begin{minipage}[b]{\linewidth}\raggedright
Estimate
\end{minipage} & \begin{minipage}[b]{\linewidth}\raggedleft
\(p_\text{Holm}\)
\end{minipage} & \begin{minipage}[b]{\linewidth}\centering
Registered
\end{minipage} & \begin{minipage}[b]{\linewidth}\centering
Sensitivity
\end{minipage} \\
\midrule\noalign{}
\endfirsthead
\toprule\noalign{}
\begin{minipage}[b]{\linewidth}\raggedright
Hypothesis
\end{minipage} & \begin{minipage}[b]{\linewidth}\raggedright
Model
\end{minipage} & \begin{minipage}[b]{\linewidth}\raggedright
Estimate
\end{minipage} & \begin{minipage}[b]{\linewidth}\raggedleft
\(p_\text{Holm}\)
\end{minipage} & \begin{minipage}[b]{\linewidth}\centering
Registered
\end{minipage} & \begin{minipage}[b]{\linewidth}\centering
Sensitivity
\end{minipage} \\
\midrule\noalign{}
\endhead
\bottomrule\noalign{}
\endlastfoot
H1 miss vs R2 (F), diff. {[}95\% CI{]} & deepseek-v4.1-flash & +0.1
{[}-1.1, +1.3{]} pp & \(p=0.002\) & yes & no (+1.9 {[}+1.2, +2.9{]}) \\
H1 miss vs R2 (F), diff. {[}95\% CI{]} & qwen3.8-flash-next & +0.4
{[}-0.7, +1.7{]} pp & \(p=0.007\) & yes & no (+2.2 {[}+1.4, +3.2{]}) \\
H1 miss vs R2 (F), diff. {[}95\% CI{]} & gemma-4-31b & -0.3 {[}-1.4,
+0.8{]} pp & \(p<0.001\) & yes & no (+1.6 {[}+0.9, +2.5{]}) \\
H2 escalated negatives (F), model vs R2 & deepseek-v4.1-flash & 206 vs
298 & \(p<0.001\) & yes & --- \\
H2 escalated negatives (F), model vs R2 & qwen3.8-flash-next & 86 vs 298
& \(p<0.001\) & yes & --- \\
H2 escalated negatives (F), model vs R2 & gemma-4-31b & 7 vs 298 &
\(p<0.001\) & yes & --- \\
H3 F recall vs exact; F+E gain & Turkish spelling of Arabic-origin name
& 69.0\% vs 100.0\%; +10 & \(p<0.001\) & yes & --- \\
H3 F recall vs exact; F+E gain & Turkish romanisation of Cyrillic name &
90.5\% vs 100.0\%; +11 & \(p<0.001\) & yes & --- \\
H4 escalated positives (F+E), model vs R2 & deepseek-v4.1-flash & 1118
vs 1099 & \(p=0.02\) & yes & no \\
H4 escalated positives (F+E), model vs R2 & qwen3.8-flash-next & 1113 vs
1099 & \(p=0.05\) & yes & no \\
H4 escalated positives (F+E), model vs R2 & gemma-4-31b & 1120 vs 1099 &
\(p=0.010\) & yes & no \\
H5 misses, minor vs compatible date & deepseek-v4.1-flash & 20/559 vs
1/558 & \(p<0.001\) & yes & yes \\
H5 misses, minor vs compatible date & qwen3.8-flash-next & 10/559 vs
15/558 & \(p=1.00\) & no & no \\
H5 misses, minor vs compatible date & gemma-4-31b & 9/559 vs 8/558 &
\(p=1.00\) & no & no \\
H6 date assessment accuracy \(>95\%\) & deepseek-v4.1-flash & 98.4\%
(lower 98.1\%) & \(p<0.001\) & yes & --- \\
H6 date assessment accuracy \(>95\%\) & qwen3.8-flash-next & 98.3\%
(lower 98.0\%) & \(p<0.001\) & yes & --- \\
H6 date assessment accuracy \(>95\%\) & gemma-4-31b & 98.5\% (lower
98.3\%) & \(p<0.001\) & yes & --- \\
\end{longtable}

As registered, every hypothesis was supported for every model except H5,
which held only for deepseek-v4.1-flash (Table 3). The safety result
does not survive the deviation. With the affected customers excluded, R2
misses no positive in pipeline F, while the models miss 1.6\% to 2.2\%;
the upper bound of the difference lies above the 2-point margin for
every model, so non-inferiority is not shown. The registered H1 result
was produced by the models escalating customers whose dates conflict
under the policy, the opposite of what the policy asked. H4 is likewise
not supported in the sensitivity analysis: the embedding-only positives
the models catch and R2 discounts are offset by positives the models
discount. H2, H3 and H6 do not depend on the affected customers.

\subsection{Why the models miss listed persons
(exploratory)}\label{why-the-models-miss-listed-persons-exploratory}

\begin{longtable}[]{@{}
  >{\raggedright\arraybackslash}p{(\linewidth - 10\tabcolsep) * \real{0.1667}}
  >{\raggedleft\arraybackslash}p{(\linewidth - 10\tabcolsep) * \real{0.1667}}
  >{\raggedleft\arraybackslash}p{(\linewidth - 10\tabcolsep) * \real{0.1667}}
  >{\raggedleft\arraybackslash}p{(\linewidth - 10\tabcolsep) * \real{0.1667}}
  >{\raggedleft\arraybackslash}p{(\linewidth - 10\tabcolsep) * \real{0.1667}}
  >{\raggedleft\arraybackslash}p{(\linewidth - 10\tabcolsep) * \real{0.1667}}@{}}
\caption{Missed positives in pipeline F, excluding the affected
customers, by the model's own assessment of the person's
entry.}\tabularnewline
\toprule\noalign{}
\begin{minipage}[b]{\linewidth}\raggedright
Model
\end{minipage} & \begin{minipage}[b]{\linewidth}\raggedleft
Positives
\end{minipage} & \begin{minipage}[b]{\linewidth}\raggedleft
Missed
\end{minipage} & \begin{minipage}[b]{\linewidth}\raggedleft
Said date conflict
\end{minipage} & \begin{minipage}[b]{\linewidth}\raggedleft
Said different name
\end{minipage} & \begin{minipage}[b]{\linewidth}\raggedleft
Discounted despite own assessment
\end{minipage} \\
\midrule\noalign{}
\endfirsthead
\toprule\noalign{}
\begin{minipage}[b]{\linewidth}\raggedright
Model
\end{minipage} & \begin{minipage}[b]{\linewidth}\raggedleft
Positives
\end{minipage} & \begin{minipage}[b]{\linewidth}\raggedleft
Missed
\end{minipage} & \begin{minipage}[b]{\linewidth}\raggedleft
Said date conflict
\end{minipage} & \begin{minipage}[b]{\linewidth}\raggedleft
Said different name
\end{minipage} & \begin{minipage}[b]{\linewidth}\raggedleft
Discounted despite own assessment
\end{minipage} \\
\midrule\noalign{}
\endhead
\bottomrule\noalign{}
\endlastfoot
deepseek-v4.1-flash & 1,097 & 21 & 18 & 3 & 0 \\
qwen3.8-flash-next & 1,097 & 24 & 6 & 1 & 17 \\
gemma-4-31b & 1,097 & 17 & 6 & 11 & 0 \\
\end{longtable}

Each model failed in its own way (Table 4). deepseek-v4.1-flash missed
almost only under minor date discrepancies (20 of 21), which it called
conflicts although the policy names them as non-conflicts; the same
failure was reported for LLM entity matching on sanctions data
\citep{smith2026opensanctions}. gemma-4-31b's misses were mostly name
judgments: it called word-order and typing variants different names.
qwen3.8-flash-next usually assessed the name as a variant and the date
as compatible and then discounted anyway, so its decision contradicted
its own rationale. A reviewer who reads only the reason would see a
correct analysis attached to a wrong decision.

\subsection{Rationale accuracy, the pilot model and
cost}\label{rationale-accuracy-the-pilot-model-and-cost}

The stated date-of-birth assessment matched the deterministic status in
98.3\% to 98.5\% of alerts (H6). The models' decisions followed their
own assessments in 98.6\% to 100.0\% of alerts. The local qwen3.8:27b,
run only in the pilot, missed 3.6\% of 112 positives and escalated
10.0\% of alerted negatives, against 13.2\% for R2 in the same pilot.

An alert cost on average 608 to 646 input and 74 to 104 output tokens,
and the full run used 15.8 million tokens across the three models.
Median latency per call was 1.4 to 9.4 seconds under twelve parallel
requests. We had registered EVREN credit as a cost measure, but the
credit header returned the same value on all 25,165 calls, so we report
tokens instead.

\section{Discussion}\label{discussion}

The comparison that matters for a bank is not ``model versus rule'' but
which part of the level-1 decision each should make. The date clause of
a sanctions policy is a deterministic comparison, and a rule makes it
without error. The models made it correctly in 98\% to 99\% of alerts,
which is accurate for a language model and too inaccurate for the step
that decides whether a listed person passes. The name clause is where a
rule is weak: a fuzzy score cannot tell a transliteration of the same
surname from a different surname, so R2 escalated every near-name
customer. That is where the models saved most of the workload.

This suggests a hybrid that we did not test: compute the date status
deterministically, let the model judge only the name, and allow a
discount on the name alone only when the model is confident and the
customer is not a near-transliteration. The failure modes we found argue
for three further controls. A model that discounts against its own
stated assessment (qwen3.8-flash-next) can be caught by a consistency
check between the structured assessments and the decision, which costs
nothing. A model that calls variants different names (gemma-4-31b) needs
those discounts sampled for review. And because stage-1 recall on
Turkish spellings of Arabic-origin names was well below that on exact
names, a bank using fuzzy screening on Turkish customers has a larger
problem before level 1 than at it.

These points bear on the AMLR requirement, applicable to EU obliged
entities from 10 July 2027, of meaningful human intervention in
automated due diligence decisions \citep[Art. 76(5)(b)]{eu2024amlr}. If
a level-1 discount is treated as such a decision, a model that reduces
workload by sending fewer alerts to humans also reduces the human
intervention in the decisions it discounts. The discounts are where the
risk sits, so they, not the escalations, are what a human review sample
should be drawn from.

Our own error is a methodological point. The construction error was in
the generator, not in the models, and the pilot was too small to make it
visible: 2 affected customers moved R2's pilot miss rate from zero to
1.8\%, which we did not investigate. A rule baseline that is error-free
by construction is a useful audit of the benchmark itself, and any miss
it makes should be traced before the confirmatory run.

\section{Limitations}\label{limitations}

The customers are synthetic, and name variants were generated by rules;
real customer names, real typing errors and real list-quality problems
will differ. The general population is far smaller relative to the watch
list than in a bank, so alert rates per customer are not representative,
and workload is reported per 1,000 screened customers of this
population. Only UK and EU lists were used, and only natural persons
with a date of birth. The policy is one level-1 policy; many banks also
use nationality, place of birth and identifiers. Each alert was judged
alone, at temperature 0, by three models of moderate size on one
domestic API with reasoning disabled; larger or reasoning models may do
better, and served model versions can change. The embedding threshold
was tuned on the pilot. The sensitivity analysis is post hoc; the
affected customers are a small, specific subset, and we did not
regenerate them, so the effect of a corrected generator on the models is
unknown.

\section{Conclusion}\label{conclusion}

Open language models that a bank can run on its own infrastructure
cleared sanctions alerts with far less level-2 workload than a rule that
applies the same policy, and they stated their reasons correctly most of
the time. They also let through about 2\% to 2\% of the listed persons
that the rule escalated, for reasons specific to each model. A
pre-registered safety test passed only because of an error in our data,
which a rule baseline exposed. On this evidence a model should not
decide the date clause, and its discounts, not its escalations, need
human review.

\section*{Declarations}\label{declarations}
\addcontentsline{toc}{section}{Declarations}

\textbf{Funding.} This research did not receive any specific grant from
funding agencies in the public, commercial, or not-for-profit sectors.

\textbf{Competing interests.} The author works for a participation bank.
The views expressed are the author's own and do not represent those of
any employer. No bank or customer data were used.

\textbf{Data and code.} The list-processing, customer-generation,
screening, disposition and analysis code, the policy, the synthetic
customers, all model outputs, the pre-registration with its SHA-256
hashes and the deviation note are available at
https://github.com/mhuyut/llm-sanctions-alert-triage. The watch-list
files themselves are not redistributed; a script downloads them from the
official sources. The UK list contains public sector information
licensed under the Open Government Licence v3.0.

\textbf{Use of AI tools.} The author directed this study: the author
selected the research topic and question, chose the experimental design
and approved its pre-registered amendments (including dropping a model
that failed the pilot gate), reviewed all of the experiment and analysis
code, monitored every experimental run and secured the computing
resources needed for all models to complete their tests, and read,
reviewed and finalized the manuscript. Claude Code (Anthropic; Claude
Opus 5.5) was used as a research assistant throughout: to draft the
design and pre-registration, write the experiment and analysis code,
trace the data-generation error reported in the paper, search and verify
the literature, and produce the LaTeX build, with all numbers generated
automatically from the analysis output. The open-weight models under
study did not contribute to the writing. The author takes full
responsibility for the content; no AI tool is listed as an author.

\renewcommand\refname{References}
\bibliography{refs.bib}

\begin{thebibliography}{34}
\providecommand{\natexlab}[1]{#1}
\providecommand{\url}[1]{\texttt{#1}}
\expandafter\ifx\csname urlstyle\endcsname\relax
  \providecommand{\doi}[1]{doi: #1}\else
  \providecommand{\doi}{doi: \begingroup \urlstyle{rm}\Url}\fi

\bibitem[Bachmann(2026)]{bachmann2026rapidfuzz}
Max Bachmann.
\newblock {RapidFuzz}: Release 3.14.6.
\newblock \url{https://github.com/rapidfuzz/RapidFuzz}, 2026.

\bibitem[{Bankac{\i}l{\i}k D{\"u}zenleme ve Denetleme
  Kurumu}(2020)]{bddk2020bs}
{Bankac{\i}l{\i}k D{\"u}zenleme ve Denetleme Kurumu}.
\newblock Bankalar{\i}n bilgi sistemleri ve elektronik bankac{\i}l{\i}k
  hizmetleri hakk{\i}nda y{\"o}netmelik.
\newblock Resm{\^i} Gazete, 15.03.2020, No. 31069, 2020.
\newblock URL
  \url{https://www.resmigazete.gov.tr/eskiler/2020/03/20200315-10.htm}.
\newblock Original text verified (Art. 25: primary and secondary systems in
  T{\"u}rkiye).

\bibitem[Bilenko et~al.(2003)Bilenko, Mooney, Cohen, Ravikumar, and
  Fienberg]{bilenko2003adaptive}
Mikhail Bilenko, Raymond Mooney, William Cohen, Pradeep Ravikumar, and Stephen
  Fienberg.
\newblock Adaptive name matching in information integration.
\newblock \emph{IEEE Intelligent Systems}, 18\penalty0 (5):\penalty0 16--23,
  2003.
\newblock \doi{10.1109/MIS.2003.1234765}.
\newblock Verified: metadata only.

\bibitem[Chen et~al.(2024)Chen, Xiao, Zhang, Luo, Lian, and Liu]{chen2024bgem3}
Jianlyu Chen, Shitao Xiao, Peitian Zhang, Kun Luo, Defu Lian, and Zheng Liu.
\newblock {M3-Embedding}: Multi-linguality, multi-functionality,
  multi-granularity text embeddings through self-knowledge distillation.
\newblock In \emph{Findings of the Association for Computational Linguistics:
  ACL 2024}, pages 2318--2335, Bangkok, Thailand, 2024. Association for
  Computational Linguistics.
\newblock \doi{10.18653/v1/2024.findings-acl.137}.

\bibitem[Chernakov et~al.(2025)Chernakov, Jafarnejad, and
  Frank]{chernakov2025adverse}
Pavel Chernakov, Sasan Jafarnejad, and Rapha{\"e}l Frank.
\newblock An agentic {LLM} framework for adverse media screening in {AML}
  compliance, 2025.
\newblock URL \url{https://arxiv.org/abs/2602.23373}.

\bibitem[Chourasia et~al.(2026)Chourasia, Kapoor, and Patil]{chourasia2026sger}
Shivam Chourasia, Hitesh Kapoor, and Nilesh Patil.
\newblock Structure-guided entity resolution: Fine-tuning {LLMs} for robust
  name matching in complex linguistic contexts.
\newblock In \emph{Proceedings of the 64th Annual Meeting of the Association
  for Computational Linguistics (Volume 6: Industry Track)}, pages 1461--1468,
  2026.
\newblock \doi{10.18653/v1/2026.acl-industry.101}.

\bibitem[Christen(2012)]{christen2012datamatching}
Peter Christen.
\newblock \emph{Data Matching: Concepts and Techniques for Record Linkage,
  Entity Resolution, and Duplicate Detection}.
\newblock Springer, Berlin, Heidelberg, 2012.
\newblock ISBN 978-3-642-31163-5.
\newblock \doi{10.1007/978-3-642-31164-2}.

\bibitem[{European Parliament and Council of the European
  Union}(2024)]{eu2024amlr}
{European Parliament and Council of the European Union}.
\newblock Regulation ({EU}) 2024/1624 of 31 may 2024 on the prevention of the
  use of the financial system for the purposes of money laundering or terrorist
  financing.
\newblock Official Journal of the European Union, L, 2024/1624, 2024.
\newblock URL \url{https://eur-lex.europa.eu/eli/reg/2024/1624/oj}.

\bibitem[{European Union}()]{eu_fsf_list}
{European Union}.
\newblock Consolidated list of persons, groups and entities subject to {EU}
  financial sanctions.
\newblock
  \url{https://data.europa.eu/data/datasets/consolidated-list-of-persons-groups-and-entities-subject-to-eu-financial-sanctions}.
\newblock Accessed 2026-11-04.

\bibitem[{FATF}(2012--2026)]{fatf2012rec}
{FATF}.
\newblock International standards on combating money laundering and the
  financing of terrorism \& proliferation: The {FATF} recommendations.
\newblock Technical report, FATF, Paris, 2012--2026.
\newblock URL
  \url{https://www.fatf-gafi.org/content/dam/fatf-gafi/recommendations/fatf-recommendations-2012.pdf}.
\newblock Adopted 16 February 2012; updated June 2026. Read via Wayback
  snapshot (2026-10-23) of the official URL.

\bibitem[Fellegi and Sunter(1969)]{fellegi1969theory}
Ivan~P. Fellegi and Alan~B. Sunter.
\newblock A theory for record linkage.
\newblock \emph{Journal of the American Statistical Association}, 64\penalty0
  (328):\penalty0 1183--1210, 1969.
\newblock \doi{10.1080/01621459.1969.10501049}.
\newblock Verified: metadata only.

\bibitem[{Foreign, Commonwealth \& Development
  Office}(2026)]{uk_sanctions_list}
{Foreign, Commonwealth \& Development Office}.
\newblock The {UK} sanctions list.
\newblock
  \url{https://www.gov.uk/government/publications/the-uk-sanctions-list}, 2026.
\newblock Data file
  \url{https://sanctionslist.fcdo.gov.uk/docs/UK-Sanctions-List.csv}. Accessed
  2026-11-04.

\bibitem[Gopal et~al.(2026)Gopal, Pravin, Kadu, Jain, and
  Krishnan]{gopal2026crossscript}
Athira Gopal, Patil~Prathamesh Pravin, Ankush Kadu, Arpit Jain, and Ashwanth
  Krishnan.
\newblock Same entity, different script: Diagnosing cross-script entity
  matching failures in language models, 2026.
\newblock URL \url{https://arxiv.org/abs/2610.31407}.

\bibitem[Gupta(2026)]{gupta2026corpfam}
Harshit Gupta.
\newblock Corporate-family resolution is not a string-matching problem: A
  public benchmark stratified by name visibility, 2026.
\newblock URL \url{https://arxiv.org/abs/2609.04269}.

\bibitem[He et~al.(2025)He, Li, Chen, Liu, Chen, Sui, Chen, Zhu, Luo, Yang, and
  Hooi]{he2025aria}
Yufei He, Ruoyu Li, Alex Chen, Yue Liu, Yulin Chen, Yuan Sui, Cheng Chen,
  Yi~Zhu, Luca Luo, Frank Yang, and Bryan Hooi.
\newblock Enabling self-improving agents to learn at test time with
  human-in-the-loop guidance.
\newblock In \emph{Proceedings of the 2025 Conference on Empirical Methods in
  Natural Language Processing: Industry Track}, pages 1625--1653, 2025.
\newblock \doi{10.18653/v1/2025.emnlp-industry.115}.

\bibitem[Holm(1979)]{holm1979simple}
Sture Holm.
\newblock A simple sequentially rejective multiple test procedure.
\newblock \emph{Scandinavian Journal of Statistics}, 6\penalty0 (2):\penalty0
  65--70, 1979.
\newblock URL \url{https://www.jstor.org/stable/4615733}.

\bibitem[Jaro(1989)]{jaro1989advances}
Matthew~A. Jaro.
\newblock Advances in record-linkage methodology as applied to matching the
  1985 census of {Tampa, Florida}.
\newblock \emph{Journal of the American Statistical Association}, 84\penalty0
  (406):\penalty0 414--420, 1989.
\newblock \doi{10.1080/01621459.1989.10478785}.
\newblock Verified: metadata only.

\bibitem[Li et~al.(2024)Li, Li, Hao, Zhang, Song, and Chen]{li2024booster}
Huahang Li, Shuangyin Li, Fei Hao, Chen~Jason Zhang, Yuanfeng Song, and Lei
  Chen.
\newblock {BoostER}: Leveraging large language models for enhancing entity
  resolution.
\newblock In \emph{Companion Proceedings of the ACM Web Conference 2024 (WWW
  '24)}, pages 1043--1046, 2024.
\newblock \doi{10.1145/3589335.3651245}.

\bibitem[Naik et~al.(2025)Naik, Dintakurthi, Hu, Wang, and
  Qiu]{naik2025coinvestigator}
Prathamesh~Vasudeo Naik, Naresh~Kumar Dintakurthi, Zhanghao Hu, Yue Wang, and
  Robby Qiu.
\newblock Co-investigator {AI}: The rise of agentic {AI} for smarter,
  trustworthy {AML} compliance narratives, 2025.
\newblock URL \url{https://arxiv.org/abs/2509.08380}.

\bibitem[Naik et~al.(2026)Naik, Dintakurthi, and Wang]{naik2026llmops}
Prathamesh~Vasudeo Naik, Naresh Dintakurthi, and Yue Wang.
\newblock Rethinking {LLMOps} for fraud and {AML}: Building a compliance-grade
  {LLM} serving stack, 2026.
\newblock URL \url{https://arxiv.org/abs/2605.11232}.

\bibitem[Naser~Eddin et~al.(2021)Naser~Eddin, Bono, Apar{\'i}cio, Polido,
  Ascens{\~a}o, Bizarro, and Ribeiro]{nasereddin2021alert}
Ahmad Naser~Eddin, Jacopo Bono, David Apar{\'i}cio, David Polido,
  Jo{\~a}o~Tiago Ascens{\~a}o, Pedro Bizarro, and Pedro Ribeiro.
\newblock Anti-money laundering alert optimization using machine learning with
  graphs, 2021.
\newblock URL \url{https://arxiv.org/abs/2112.07508}.

\bibitem[Newcombe(1998)]{newcombe1998paired}
Robert~G. Newcombe.
\newblock Improved confidence intervals for the difference between binomial
  proportions based on paired data.
\newblock \emph{Statistics in Medicine}, 17\penalty0 (22):\penalty0 2635--2650,
  1998.
\newblock
  \doi{10.1002/(SICI)1097-0258(19981130)17:22<2635::AID-SIM954>3.0.CO;2-C}.

\bibitem[{Office of Foreign Assets Control, U.S. Department of the
  Treasury}()]{ofac_sdn}
{Office of Foreign Assets Control, U.S. Department of the Treasury}.
\newblock Specially designated nationals and blocked persons list ({SDN}).
\newblock \url{https://sanctionslist.ofac.treas.gov/Home/SdnList}.
\newblock Accessed 2026-11-04.

\bibitem[Pouliquen et~al.(2005)Pouliquen, Steinberger, Ignat, Temnikova, and
  Widiger]{pouliquen2005multilingual}
Bruno Pouliquen, Ralf Steinberger, Camelia Ignat, Irina Temnikova, and Anna
  Widiger.
\newblock Multilingual person name recognition and transliteration.
\newblock \emph{Corela}, \penalty0 (HS-2), 2005.
\newblock \doi{10.4000/corela.1219}.
\newblock Verified: metadata only.

\bibitem[Smith et~al.(2026)Smith, Sesodia, Lindenberg, and Schroeder~de
  Witt]{smith2026opensanctions}
Chandler Smith, Magnus Sesodia, Friedrich Lindenberg, and Christian
  Schroeder~de Witt.
\newblock {OpenSanctions} pairs: Large-scale entity matching with {LLMs}, 2026.
\newblock URL \url{https://arxiv.org/abs/2603.11051}.

\bibitem[S{\o}gaard et~al.(2023)S{\o}gaard, Hershcovich, and
  de~Lhoneux]{sogaard2023prereg}
Anders S{\o}gaard, Daniel Hershcovich, and Miryam de~Lhoneux.
\newblock A two-sided discussion of preregistration of {NLP} research.
\newblock In \emph{Proceedings of the 17th Conference of the European Chapter
  of the Association for Computational Linguistics}, pages 83--93, 2023.
\newblock \doi{10.18653/v1/2023.eacl-main.6}.

\bibitem[Torres et~al.(2026)Torres, Cheng, and Hu]{torres2026triage}
Dorothy Torres, Wei Cheng, and Ke~Hu.
\newblock Explainable {AML} triage with {LLMs}: Evidence retrieval and
  counterfactual checks, 2026.
\newblock URL \url{https://arxiv.org/abs/2604.19755}.

\bibitem[{T{\"u}rkiye B{\"u}y{\"u}k Millet Meclisi}(2013)]{tr6415}
{T{\"u}rkiye B{\"u}y{\"u}k Millet Meclisi}.
\newblock 6415 say{\i}l{\i} ter{\"o}rizmin finansman{\i}n{\i}n {\"o}nlenmesi
  hakk{\i}nda kanun.
\newblock Resm{\^i} Gazete, 16.02.2013, No. 28561, 2013.
\newblock URL \url{https://www.mevzuat.gov.tr/MevzuatMetin/1.5.6415.pdf}.

\bibitem[{T{\"u}rkiye B{\"u}y{\"u}k Millet Meclisi}(2016)]{tr6698}
{T{\"u}rkiye B{\"u}y{\"u}k Millet Meclisi}.
\newblock 6698 say{\i}l{\i} ki{\c{s}}isel verilerin korunmas{\i} kanunu.
\newblock Resm{\^i} Gazete, 07.04.2016, No. 29677, 2016.
\newblock URL \url{https://www.mevzuat.gov.tr/MevzuatMetin/1.5.6698.pdf}.
\newblock Art. 9 amended by Law 7499 (02.03.2024).

\bibitem[{T{\"u}rkiye B{\"u}y{\"u}k Millet Meclisi}(2020)]{tr7262}
{T{\"u}rkiye B{\"u}y{\"u}k Millet Meclisi}.
\newblock 7262 say{\i}l{\i} kitle {\.i}mha silahlar{\i}n{\i}n
  yay{\i}lmas{\i}n{\i}n finansman{\i}n{\i}n {\"o}nlenmesine {\.i}li{\c{s}}kin
  kanun.
\newblock Resm{\^i} Gazete, 31.12.2020, No. 31351 (5. M{\"u}kerrer), 2020.
\newblock URL \url{https://www.mevzuat.gov.tr/MevzuatMetin/1.5.7262.pdf}.

\bibitem[van Miltenburg et~al.(2021)van Miltenburg, van~der Lee, and
  Krahmer]{vanmiltenburg2021prereg}
Emiel van Miltenburg, Chris van~der Lee, and Emiel Krahmer.
\newblock Preregistering {NLP} research.
\newblock In \emph{Proceedings of the 2021 Conference of the North American
  Chapter of the Association for Computational Linguistics: Human Language
  Technologies}, pages 613--623, 2021.
\newblock \doi{10.18653/v1/2021.naacl-main.51}.

\bibitem[Wahib et~al.(2026)Wahib, Tran, Mourad, Muti, and
  Dvornik]{wahib2026spectra}
Miray Wahib, Ethan Tran, Rea Mourad, Mira Muti, and Nikita Dvornik.
\newblock Spectra: A rules-driven {LLM} pipeline for automated {KYC} document
  processing, 2026.
\newblock URL \url{https://arxiv.org/abs/2609.25474}.

\bibitem[Wilson(1927)]{wilson1927probable}
Edwin~B. Wilson.
\newblock Probable inference, the law of succession, and statistical inference.
\newblock \emph{Journal of the American Statistical Association}, 22\penalty0
  (158):\penalty0 209--212, 1927.
\newblock \doi{10.1080/01621459.1927.10502953}.

\bibitem[Zeakis et~al.(2026)Zeakis, Papadakis, Skoutas, and
  Koubarakis]{zeakis2026distiller}
Alexandros Zeakis, George Papadakis, Dimitrios Skoutas, and Manolis Koubarakis.
\newblock {DistillER}: Knowledge distillation in entity resolution with large
  language models, 2026.
\newblock URL \url{https://arxiv.org/abs/2602.05452}.

\end{thebibliography}

\end{document}